\documentclass[letterpaper]{article} 
\usepackage{aaai2026}  
\usepackage{times}  
\usepackage{helvet}  
\usepackage{courier}  
\usepackage[hyphens]{url}  
\usepackage{graphicx} 
\usepackage{natbib}  
\usepackage{caption} 
\usepackage{algorithm}
\usepackage{algorithmic}

\usepackage{newfloat}
\usepackage{listings}
\DeclareCaptionStyle{ruled}{labelfont=normalfont,labelsep=colon,strut=off} 
\floatstyle{ruled}
\newfloat{listing}{tb}{lst}{}
\floatname{listing}{Listing}
\usepackage{amsmath}
\usepackage{amsfonts}
\usepackage{amsmath}
\usepackage{amsfonts}
\usepackage{pifont}
\usepackage[table]{xcolor}
\usepackage{booktabs}
\usepackage{multirow}
\usepackage{makecell}
\usepackage{threeparttable}
\usepackage{tabularx}
\usepackage{cuted}
\usepackage{capt-of}
\usepackage[toc,page]{appendix}
\usepackage{subcaption}
\usepackage{lipsum}
\usepackage{placeins}
\title{RefineVAD: Semantic-Guided Feature Recalibration \\for Weakly Supervised Video Anomaly Detection}

\author{
    Junhee Lee\equalcontrib\textsuperscript{\rm 1},
    ChaeBeen Bang\equalcontrib\textsuperscript{\rm 1},
    MyoungChul Kim\equalcontrib\textsuperscript{\rm 2},
    MyeongAh Cho\thanks{Corresponding author.}\textsuperscript{\rm 2}
}

\affiliations{
    \textsuperscript{\rm 1}Department of Artificial Intelligence, Kyung Hee University\\
    \textsuperscript{\rm 2}Department of Software Convergence, Kyung Hee University\\
    \{jhlee39, coqls1229, 2018102095, maycho\}@khu.ac.kr
}

\begin{document}

\maketitle

\begin{figure*}[t]
  \centering
  \includegraphics[width=0.9\textwidth]{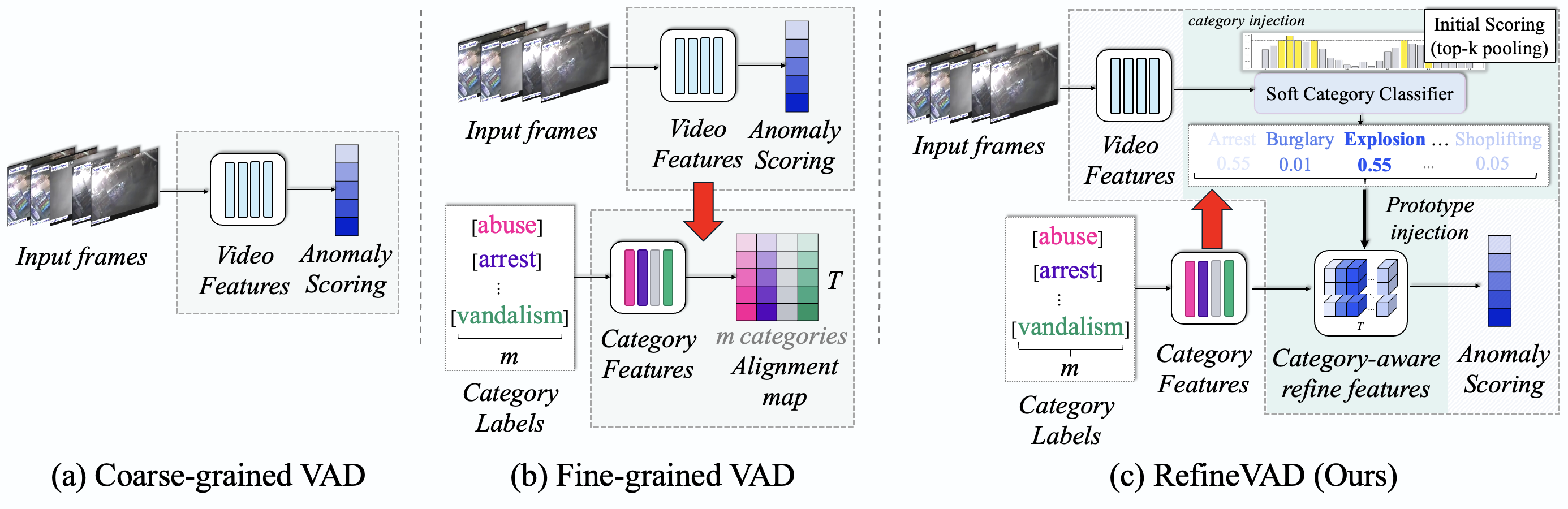}
  \caption{
    Comparison of WVAD paradigms. 
    (a) Coarse-grained model predicts anomaly scores per snippet. 
    (b) Fine-grained method introduces auxiliary video-level category classification, but these labels are not utilized during anomaly scoring. 
    (c) RefineVAD (Ours): category-aware soft classification guides feature enhancement with learnable prototypes.
  }
  \label{fig1}
\end{figure*}

\begin{abstract}
Weakly-Supervised Video Anomaly Detection aims to identify anomalous events using only video-level labels, balancing annotation efficiency with practical applicability. However, existing methods often oversimplify the anomaly space by treating all abnormal events as a single category, overlooking the diverse semantic and temporal characteristics intrinsic to real-world anomalies. Inspired by how humans perceive anomalies, by jointly interpreting temporal motion patterns and semantic structures underlying different anomaly types, we propose RefineVAD, a novel framework that mimics this dual-process reasoning. Our framework integrates two core modules. The first, Motion-aware Temporal Attention and Recalibration (MoTAR), estimates motion salience and dynamically adjusts temporal focus via shift-based attention and global Transformer-based modeling. The second, Category-Oriented Refinement (CORE), injects soft anomaly category priors into the representation space by aligning segment-level features with learnable category prototypes through cross-attention. By jointly leveraging temporal dynamics and semantic structure, explicitly models both ``how'' motion evolves and ``what'' semantic category it resembles. Extensive experiments on WVAD benchmark validate the effectiveness of RefineVAD and highlight the importance of integrating semantic context to guide feature refinement toward anomaly-relevant patterns.
\end{abstract}


\begin{links}
    \link{Code}{https://github.com/VisualScienceLab-KHU/RefineVAD}
\end{links}

\section{Introduction}

Video Anomaly Detection (VAD) aims to identify unexpected or abnormal events within video streams and has become increasingly important in applications such as surveillance, public safety, and industrial monitoring. While fully supervised methods achieve precise anomaly localization, they require costly frame-level annotations, limiting their scalability in real-world scenarios. Unsupervised approaches, which learn solely from normal patterns, reduce labeling costs but often suffer from poor generalization and high false-positive rates. To balance annotation efficiency with detection performance, Weakly Supervised Video Anomaly Detection (WVAD) has emerged as a compelling alternative. WVAD methods rely exclusively on video-level labels—indicating whether a video contains any anomaly—and typically adopt a Multiple Instance Learning (MIL) paradigm, wherein each video is treated as a bag of segments, and at least one segment in an anomalous video is assumed to contain abnormal content.

However, recognizing abnormal events in videos requires more than coarse video-level classification. Human perception leverages two complementary dimensions when detecting anomalies: \textbf{(1) contextual motion dynamics that evolve over time}—such as abrupt or irregular movements—and \textbf{(2) prior knowledge of what type of anomaly is occurring}, allowing us to interpret visual patterns semantically. This motivates a hierarchical reasoning framework that first attends to how motion unfolds and then reasons about what kind of abnormal event it resembles. Such reasoning enables more robust and interpretable anomaly detection by aligning perceptual cues with semantic understanding.

Despite recent advancements, existing WVAD methods exhibit two critical limitations that impede this hierarchical reasoning. First, \textbf{temporal modeling is often shallow or inflexible}, relying on fixed pooling operations or simple aggregation schemes that fail to adapt to the diverse motion characteristics of real-world anomalies. Since many anomalies are defined by dynamic, non-uniform, or context-specific motion patterns, this temporal rigidity significantly hampers localization accuracy. Second, and more fundamentally, most WVAD frameworks \textbf{treat all abnormal events as a single generic class}, ignoring the semantic diversity across anomaly types. While the original VAD task framed anomaly detection as a binary classification problem—determining whether each snippet is normal or abnormal, in practice, anomalies exhibit highly distinctive visual and motion patterns. For instance, fighting involves abrupt, bidirectional motion, while explosions are characterized by sudden flashes and spatial bursts. Ignoring such distinctions limits the model's ability to capture discriminative features, thereby reducing both interpretability and localization performance. Although some recent works have attempted to incorporate anomaly categories, these are typically used only for auxiliary video-level classification and not integrated into the anomaly scoring or representation learning processes. Consequently, semantic distinctions remain unexploited during temporal localization, and the model fails to learn category-aware features crucial for precise detection.

To address these challenges, we propose \textbf{RefineVAD}, a novel framework that introduces \textbf{motion-aware temporal recalibration} and \textbf{category-conditioned refinement} into the anomaly detection process. RefineVAD integrates two key components—\textbf{MoTAR} and \textbf{CORE}—to explicitly models both \textit{``how''} motion evolves and \textit{``what''} semantic category it resembles, enabling structured anomaly reasoning under weak supervision.
The Motion-aware Temporal Attention and Recalibration (MoTAR) module captures motion-intensity-aware relationships between segments by dynamically shifting feature channels in proportion to estimated motion salience. It further employs a lightweight Transformer-based global attention mechanism to model long-range dependencies, allowing the model to focus on critical temporal regions without relying on rigid windowing or fixed aggregation.

The Category-Oriented Refinement (CORE) module incorporates semantic priors by selectively injecting soft category prototypes into the representation space. It first predicts a video-level anomaly category distribution via soft classification, which is then used to compute a weighted combination of learnable category embeddings. Each embedding serves as a prototype encoding the characteristic patterns of a specific anomaly type. These category-aware combined embeddings are injected into the snippet-level features via cross-attention, guiding the model to focus on temporally and semantically meaningful regions during scoring.
By combining these two components, RefineVAD effectively disentangles and recomposes multi-modal evidence to enhance anomaly localization. It captures the temporal evolution of motion and aligns it with category-relevant semantics, thereby improving both detection precision and interpretability.

Our contributions are summarized as follows:

\begin{itemize}
    \item We propose RefineVAD, a novel WVAD framework that jointly models temporal and semantic contexts for anomaly localization under weak supervision.
    \item We design MoTAR, a motion-aware temporal attention module that adaptively recalibrates temporal feature flows based on motion salience.
    \item We introduce CORE, a category-oriented refinement module that injects soft category priors into snippet-level representations via attention-guided semantic alignment.
    \item We demonstrate that RefineVAD significantly outperforms previous state-of-the-art methods on multiple benchmark datasets through comprehensive ablation studies.
\end{itemize}

\section{Related Works}
\paragraph{Weakly Supervised Video Anomaly Detection. }
VAD aims to identify events that deviate from typical or expected normal patterns in video streams. Traditional fully supervised learning methods detect anomalies using frame-level annotations, which are expensive and difficult to apply at scale ~\cite{zhu2021video, majhi2021weakly, majhi2021dam}. On the other hand, Unsupervised VAD (UVAD) methods trained only on normal data struggle to robustly characterize normality and lead to reconstruction and prediction errors ~\cite{hasan2016learning, liu2018future, wu2019deep, cai2021appearance}. To bridge this gap, WVAD has emerged and significantly reduce the annotation cost since it only requires video-level labels to indicate whether anomalous signs are present in the video ~\cite{wu2020not, zaheer2020claws, tian2021weakly, chen2023tevad,  cho2023look, wan2020weakly}. Most WVAD methods are based on the Multiple-Instance Learning (MIL) framework to utilize anomalous video labels without frame annotations. A foundational approach applies a ranking loss to enforce higher scores for anomalous snippets compared to normal ones~\cite{sultani2018real}. Subsequent works build upon this paradigm by introducing techniques such as two-stage pseudo-labeling or surrogate optimization~\cite{feng2021mist}. These MIL methods typically follow a pipeline that extracts pre-trained features (e.g., C3D~\cite{tran2015learning}/I3D~\cite{carreira2017quo}) and then trains a snippet classifier.

\begin{figure*}[t]
\centering
\includegraphics[width=1.0\textwidth]{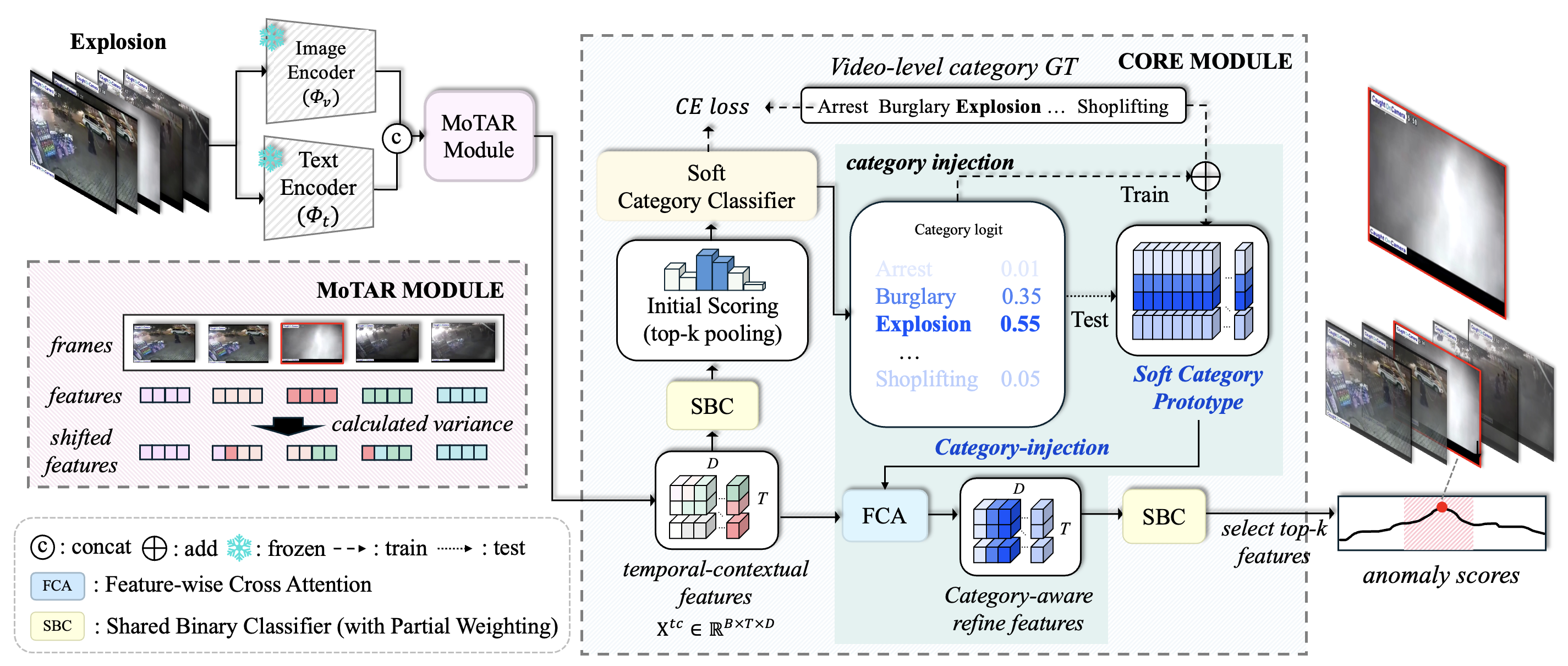} 
\caption{The overall architecture of proposed RefineVAD. Visual and textual features are extracted from each segment and fused before being processed by \textbf{MoTAR}, which adaptively recalibrates temporal features based on motion salience. The resulting features are then refined by \textbf{CORE}, which injects soft category priors via cross-attention with learnable prototypes, enabling category-aware anomaly localization.}
\label{fig2}
\end{figure*}

\paragraph{Category and Semantic-Guided VAD. }
More recent approaches shift towards leveraging semantic priors and category awareness to enhance weakly supervised video anomaly detection.
VadCLIP~\cite{wu2024vadclip} applies frozen CLIP\cite{radford2021clip} to WSVAD via a dual-branch design: one branch handles binary classification, while the other aligns frame features with text label embeddings. This enables both coarse- and fine-grained anomaly detection.
TPWNG~\cite{yang2024text} leverages CLIP with learnable prompts and event categories as text inputs, generating snippet-level pseudo-labels and training a video classifier in a self-learning loop with temporal context modeling.
PEMIL~\cite{chen2024prompt} formulates WSVAD as prompt learning, injecting semantic priors via anomaly and normal context prompts constructed from class label words.
While these methods effectively exploit CLIP’s vision-language capability, they rely on discrete category labels or handcrafted prompts, which may limit adaptability to ambiguous or overlapping anomalies. Moreover, they often overlook inter-category dependencies or the channel-wise nature of visual representations.

\section{Proposed Methods}

To address the limitations of existing WVAD methods—namely, their static temporal modeling and disregard for semantic category structures—we propose RefineVAD, a modular framework that introduces contextual awareness at two complementary levels: \textbf{(1) motion-aware temporal recalibration}, and \textbf{(2) category-conditioned semantic refinement}. Rather than relying solely on visual embeddings or treating all anomalies as a single generic class, RefineVAD leverages both structured visual and textual information, dynamically models motion salience over time, and integrates category-aware priors to enhance anomaly representation and scoring.

As illustrated in Figure~\ref{fig2}, RefineVAD follows the MIL setup by dividing each input video—annotated with a video-level anomaly label—into $T$ fixed-length segments. Each segment is independently encoded using pretrained visual and textual encoders to extract modality-specific features. These features are then concatenated to form a joint multi-modal representation for each segment, capturing both appearance and semantic context. This joint representation is first processed by the MoTAR (Motion-aware Temporal Attention and Recalibration) module. MoTAR dynamically adjusts feature channel shift ratios based on segment motion intensity to achieve adaptive temporal recalibration based on the degree of motion irregularity. In parallel, it captures long-range temporal dependencies using a lightweight Transformer-based attention mechanism~\cite{vaswani2017attention}, allowing flexible and adaptive modeling of without rigid pooling or fixed windowing.

Next, the motion-contextual features are refined by the CORE (Category-Oriented Refinement) module, which incorporates semantic priors into the representation space. CORE first performs a soft classification at the video-level to estimate the probability distribution over anomaly categories. These probabilities are used to compute a weighted combination of learnable category prototypes, each encoding representative patterns of a specific anomaly type. The resulting soft category embedding is then injected into all segment-level features via cross-attention, enabling the model to emphasize segments that are not only temporally salient but also semantically consistent with the predicted anomaly type. 

By jointly modeling motion intensity and category relevance, RefineVAD enables the disentanglement and recomposition of multi-modal cues to enhance anomaly localization. The final category-aware, temporally contextualized features are passed through a lightweight classifier to compute segment-level anomaly scores, which are subsequently aggregated to produce the video-level anomaly prediction.

\subsection{Motion-aware Temporal Attention Recalibration}

\begin{figure}[t]
\centering
\includegraphics[width=1.0\columnwidth]{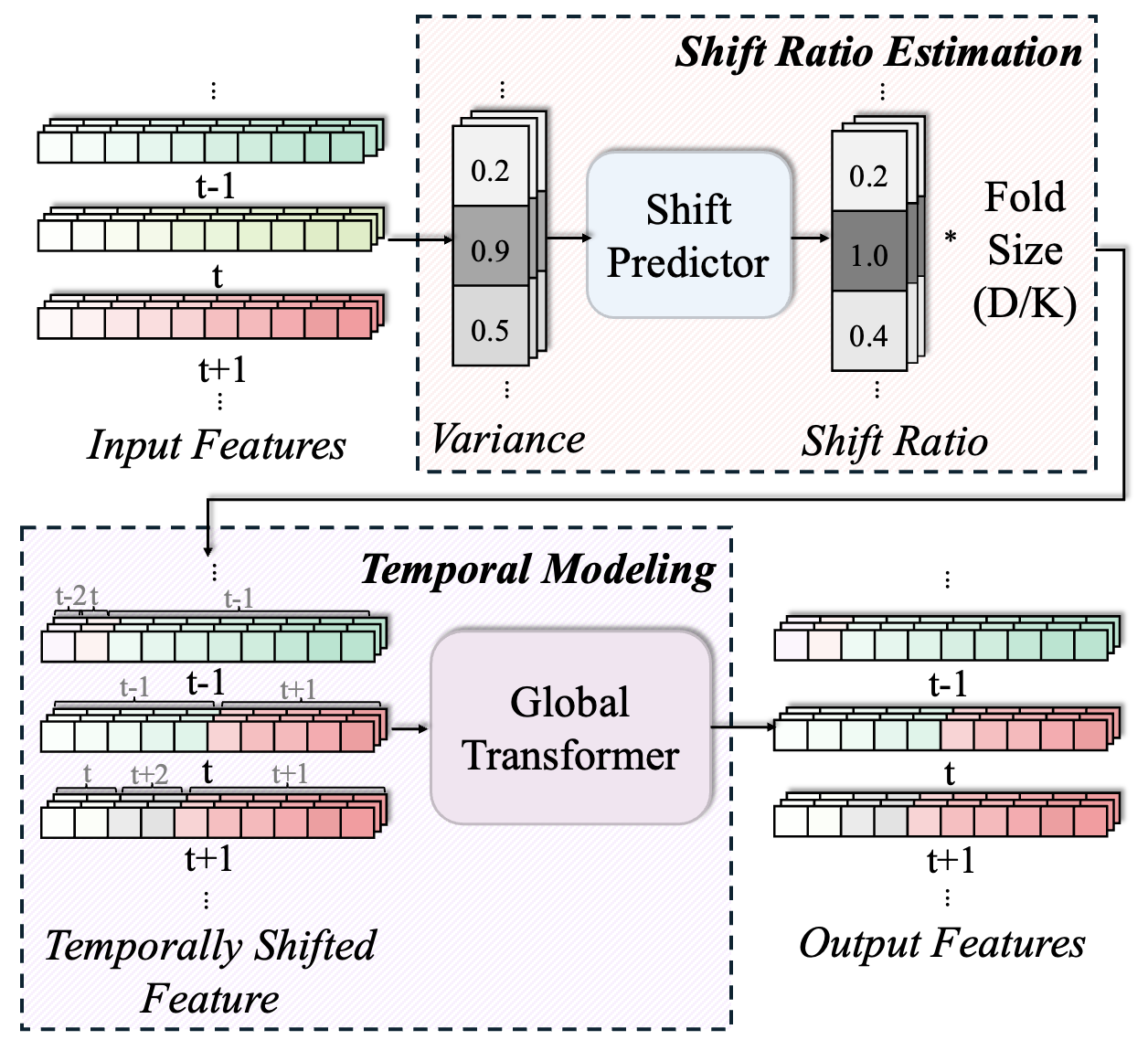} 
\caption{Overview of MoTAR. Motion variance guides adaptive channel shifting, enabling dynamic temporal feature aggregation. A Global Transformer captures long-range dependencies, improving sensitivity to diverse motions.}
\label{fig3}
\end{figure}

Anomalous events in videos often exhibit irregular motion dynamics—some are abrupt, while others remain deceptively static. To accommodate this variability, we propose MoTAR, a lightweight and effective temporal feature modeling module.

MoTAR is built upon the Temporal Shift Module (TSM)~\cite{lin2019tsm}, which introduces temporal interactions by shifting a portion of feature channels forward and backward in time. While TSM is widely used in video understanding due to its efficiency and parameter free design, it applies a fixed shift ratio across all frames, limiting its adaptability to frames with varying motion intensities.

To address this, MoTAR first estimates the motion intensity per-frame using feature differences. As shown in Figure~\ref{fig3}, given an input from the vision and text encoder $\mathbf{X} = [\mathbf{x}_1, \dots, \mathbf{x}_T] \in \mathbb{R}^{T \times D}$, where $T$ is the number of sequence and $\mathbf{D}$ is the feature dimension, we compute $\Delta_t = \mathbf{x}_t - \mathbf{x}_{t-1}$ and obtain $\mathbf{v_t} = \mathrm{Var}(\Delta_t)$
where $\mathbf{v_t}$ reflects local motion intensity. A higher variance indicates more significant motion, suggesting the need for broader temporal context aggregation. This variance-based measure is representation-agnostic, noise-robust, and enables adaptive temporal shifting with minimal computational overhead.

The variance vector is processed by a lightweight MLP-based shift predictor to obtain a shift ratio $r_t \in [0, 1]$:
\begin{equation}
    r_t = \sigma(W_3 \cdot \phi(W_2 \cdot \phi(W_1 v_t)))
\end{equation}
where $\phi$ denotes the GELU activation and $\sigma$ is the sigmoid function.The number of channels to shift across features is given by:
\begin{equation}
    s_t = \left\lfloor r_t \cdot \frac{D}{K} \right\rfloor
\end{equation}
where $K$ is a predefined folding factor. The temporally shifted output of the $t$-th segment is then constructed as:
\begin{equation}
    \mathbf{y}_t = 
    \begin{bmatrix}
        \mathbf{x}_{t-1}^{(1:s_t)} &
        \mathbf{x}_{t+1}^{(s_t:2s_t)} &
        \mathbf{x}_t^{(2s_t:D)}
    \end{bmatrix}
\end{equation}
This dynamic construction allows the model to allocate more shifted channels to high-motion frames, while preserving static content in low-variance frames.

We further apply a lightweight Transformer encoder over the shifted sequence to incorporate long-range temporal dependencies, producing temporal-contextual features.:
\begin{equation}
    \mathbf{X}^{\text{tc}} = \mathrm{Transformer}([\mathbf{y}_1, \dots, \mathbf{y}_T])
\end{equation}

In summary, MoTAR extracts temporal-contextual feature alignments through adaptive shifting and capturing global temporal context. Its parameter-free shifting ensures computational efficiency, making it suitable for real-time or large-scale video anomaly detection applications.

Detailed numerical results and a comprehensive analysis of the computational efficiency are provided in Appendix F.

\subsection{Category-Oriented Refinement}

To enrich temporal contextual feature with semantic information, we introduce CORE module, a key component of the RefineVAD framework. CORE systematically injects category‑specific semantics into segment-level features, thereby guiding the model to better focus on distinctive aspects of each abnormal event through semantic reasoning. The module consists of two main stages. \textbf{(1) Soft Category Classification}: We first distinguish the anomaly category associated with a given video sample by applying a soft classification mechanism, which estimates the probability of the sample belonging to each predefined anomaly class. \textbf{(2) Category Prototype Injection}: These probabilities are then used as weights to softly combine category-specific prototype embeddings, resulting in a unified semantic vector that captures both dominant and complementary category traits. This vector is further injected into individual temporal features via a cross‑attention mechanism, producing category‑aware refined features that dynamically align each segment with the most relevant category semantics. By enriching the representation with this semantically guided context, the model achieves improved discriminability in the final stage of anomaly scoring.

\noindent\textbf{Soft Category Classification. }Let $\mathbf{X}^{\text{tc}} = [\mathbf{x}_1, \dots, \mathbf{x}_T]$ be the temporal-contextual feature sequence generated by the MoTAR module, where each $\mathbf{x}_t^{\text{tc}} \in \mathbb{R}^{D}$ corresponds to the contextualized feature of the $t$-th segment. Each segment-level feature $\mathbf{x}_t^{\text{tc}}$ is processed by relatively simple fully connected (FC) layers consisting of dropout and batch normalization layers for scoring. These roughly estimated anomaly scores are converted into normalized weights which are used to aggregate the segment features along the temporal dimension. Since anomalous regions are relatively sparse within a video, it is more advantageous to focus on high-confidence segments rather than the entire video when distinguishing between categories. 

The pooled video feature is passed through a soft category classifier, yielding logits that we reshape into a matrix $\mathbf{z}\in\mathbb{R}^{\mathbf C\times 2}$, where $\mathbf{C}$ is the number of anomaly categories present in the dataset. Each row of this matrix corresponds to the ‘normal’ and ‘abnormal’ scores for a given category. For each category $c$, we denote the corresponding logit scores for the “normal” and “abnormal” classes as:

\begin{equation}
    \mathbf{z}_c = 
    \begin{bmatrix}
    z_c^n \ z_c^a
    \end{bmatrix}
    \;\in\;\mathbb{R}^2.
\end{equation}

\noindent For each category $c$, we apply a normal vs.\ anomaly softmax per category to obtain the anomaly probability $p_c^a$. This type of separation into anomaly and normal cases is motivated by the observation that normal samples encompass a highly diverse range of conditions, making it difficult for the model to learn a coherent representation if they are treated as a single additional anomaly category. Unlike earlier works that risked treating "normal" as just another anomaly type, we define the normal state by the absence of strong anomaly-specific characteristics, thereby avoiding distortion of learned representations. This is precisely why we designed the model to output the logits in this manner. 

The category-wise anomaly probabilities serve as modulation signals that guide how the input representation is refined, enabling the model to softly encode the relevance of different anomaly categories based on their activation levels. Instead of selecting only the most likely category, as is typical in standard classification, the model derives a set of importance weights from the full anomaly probability distribution, reflecting its confidence across all categories. 

\noindent Formally, we define the weights of each category as:
\begin{equation}
w_c =
\begin{cases}
\dfrac{1}{C}, 
& \text{if } S = \emptyset,\\[1.2em]
\dfrac{\exp\bigl(z_c^a\bigr)}{\displaystyle\sum_{k=1}^C \exp\bigl(z_k^a\bigr)},
& \text{if } S \neq \emptyset.
\end{cases}
\end{equation}
where $S = \{\,c \mid p_c^a \ge \tau\,\}$ is the set of categories whose anomaly probability exceeds a threshold $\tau$. This allows the model to naturally separate normal inputs from anomaly categories without introducing a dedicated "normal" prototype, enabling more faithful and effective learning that better reflects how anomalies are identified in real-world settings.

\noindent\textbf{Category Prototype Injection. }These weights are then used to softly combine the corresponding category embeddings, which act as prototypes representing each category, resulting in a feature representation that captures both shared and distinctive anomaly characteristics. Prototypes are learnable and arranged row-wise in the matrix:
\begin{equation}
\mathbf{E}
=
\begin{bmatrix}
\mathbf{e}_1^\mathsf{T}\\
\vdots\\
\mathbf{e}_C^\mathsf{T}
\end{bmatrix}
\;\in\;\mathbb{R}^{C\times d_{\mathrm{emb}}}.
\end{equation}

\noindent
To encode contextual category semantics, we compute a soft category embedding vector $\mathbf{v}$ as a weighted sum over learnable category embeddings:
\begin{equation}
\mathbf{v} = \sum_{c=1}^{C} w_c \, \mathbf{e}_c,
\end{equation}
where $w_c$ represents the model's estimated anomaly probability for the $c$-th category, and $\mathbf{e}_c$ is its corresponding embedding. This formulation allows the model to softly blend semantic cues across categories, capturing both confident and ambiguous class-related anomaly traits. During training, we enhance the soft category embedding $\mathbf{v}$ by adding the ground-truth class embedding, resulting in $\mathbf{v}_{\text{train}} = \mathbf{v} + \mathbf{e}_{y}$. This encourages class-aware specialization in the learned representations.

The embedding $\mathbf{v}$ is incorporated into the temporal-contextual features via a multi-head cross-attention mechanism. Specifically, $\mathbf{v}$ serves as the global query, while the temporal-contextual features act as both keys and values. This operation refines each segment feature $\mathbf{x}_t^{\text{tc}}$ into a category-aware refine feature $\mathbf{x}_t^\text{ca}$:
\begin{equation}
\mathbf{x}_t^{\text{ca}} = \text{CrossAttn}(\mathbf{v}, \mathbf{x}_t^{\text{tc}}, \mathbf{x}_t^{\text{tc}}),
\end{equation}
effectively aligning low-level feature patterns with high-level category cues.

Finally, anomaly scoring is performed on the aligned features described above. This scoring part shares parameters with the earlier rough scoring function that was used for pooling features to be passed into the category classifier, enabling some degree of mutual learning during training.

We combine a Top‑$k$ MIL ranking loss \cite{sultani2018real} $\mathcal{L}_{\mathrm{MIL}}$ to promote truly anomalous segments over normal ones, and an improved GMM‑based smoothing loss \cite{wang2025learning} $\mathcal{L}_{\mathrm{GMM}}$.  The original method fitted a Gaussian mixture by computing per‑video mean and variance across all segments; we enhance it by injecting our weighted category embedding into the mixture to more clearly reflect category‑specific traits.  We also use a category classification loss $\mathcal{L}_{\mathrm{cat}}$, where Eq.~5 outputs a two-logit vector $z_{c}$ for each category $c$. For a video with ground-truth category $c^{*}$, we assign the target $t_{c}=[0,1]$ if $c=c^{*}$ and $t_{c}=[1,0]$ otherwise, and define
$\mathcal{L}_{\mathrm{cat}}
=
\sum_{c=1}^{C} \mathrm{BCE}_{2}\!\left(\sigma(z_{c}), t_{c}; w_{c}\right)$,
with optional class weights $w_{c}$ to mitigate imbalance. This encourages the semantic embeddings to remain category-discriminative while retaining anomaly relevance.

The overall training objective is $\mathcal{L}_{\mathrm{total}}
=
\mathcal{L}_{\mathrm{MIL}}
+ \lambda_{1}\,\mathcal{L}_{\mathrm{GMM}}
+ \lambda_{2}\,\mathcal{L}_{\mathrm{cat}}$.

\section{Experiments}

\subsection{Experimental Setup}

We evaluate our method on two widely used benchmark datasets for WSVAD: UCF-Crime\cite{sultani2018real} and XD-Violence\cite{wu2020not}. \textbf{UCF-Crime}. For evaluation, we follow standard protocols. On UCF-Crime, we use the Area Under the Curve (AUC) and AUC for anomaly videos (AnoAUC) at the frame level. On XD-Violence, we use frame-level Average Precision (AP) and report mean Average Precision (mAP) across IoU thresholds from 0.1 to 0.5, along with the average score (AVG) for segment-level performance.

\begin{table}[t]
\centering
\setlength{\tabcolsep}{1mm}
{
\small
\resizebox{0.48\textwidth}{!}{
\begin{tabular}{clccc}
\toprule
Approach & Method & Source& \makecell{UCF-\\AUC(\%)} & \makecell{XD-\\AP(\%)} \\
\midrule
\multirow{3}{*}{\makecell[c]{semi\\-sup.}}         & SVM baseline &-& 50.10 & 50.80 \\
     & OCSVM~\citeyearpar{scholkopf1999support}  &NeurIPS& 63.20 & 28.63 \\
         & Hansan et al.~\citeyearpar{hasan2016learning} &CVPR& 51.20 & 31.25 \\
\midrule
\multirow{13}{*}{\makecell[c]{weakly\\-sup.}}          &            Ju et al.~\citeyearpar{ju2022prompting} &ECCV& 84.72 & 76.57 \\
         & Sultani et al.~\citeyearpar{sultani2018real}&CVPR& 84.14 & 75.18 \\
         & Wu et al.~\citeyearpar{wu2020not}&ECCV& 84.57 & 80.00 \\
         & AVVD~\citeyearpar{wu2022weakly}& TMM& 82.45 & 78.10 \\
     & RTFM~\citeyearpar{tian2021weakly} &ICCV& 85.66 & 78.27 \\
         & DMU~\citeyearpar{zhou2023dual} & AAAI& 86.75 & 81.66 \\
         & UMIL~\citeyearpar{lv2023unbiased} &CVPR& 86.75 & - \\
         & CLIP-TSA~\citeyearpar{joo2023clip} &ICIP& 87.58 & 82.17 \\
         & HSN~\citeyearpar{majhi2024human} & CVIU&85.45 & -\\
         &IFS-VAD~\citeyearpar{zhong2024inter} &TCSVT &85.47& 83.14\\
         & VadCLIP~\citeyearpar{wu2024vadclip} & AAAI & 88.02 & 84.57 \\
         &PEMIL~\citeyearpar{chen2024prompt} & CVPR&86.83& \underline{88.21} \\
         &ReFLIP~\citeyearpar{dev2024reflip} & TCSVT&88.57& 85.81 \\
         &CMHKF~\citeyearpar{wang2025cmhkf} & ACL &-& 86.57 \\
         &Ex-VAD~\citeyearpar{huangex} & ICML&88.29 & 86.52 \\
         &$\pi$-VAD~\citeyearpar{majhi2025just} & CVPR&\textbf{90.33}& 85.37\\
         & \textbf{RefineVAD (Ours)} && \underline{88.92} & \textbf{88.66} \\
         
\bottomrule
\end{tabular}
}
\vspace{1mm}
\caption{\fontsize{10}{12}\selectfont Comparison of state-of-the-art methods on UCF-Crime and XD-Violence datasets. Best result is \textbf{bolded} and second best result is
\underline{underlined}.}
\label{tab:comparision_of_sota}
}
\end{table}

\begin{table}[t]
\centering
\setlength{\tabcolsep}{1mm} 
\renewcommand{\arraystretch}{1.1}
{
\small
\begin{tabular}{lcccccc}
\toprule
\multirow{2}{*}{Method} & \multicolumn{6}{c}{mAP@IOU (\%)} \\
 & 0.1 & 0.2 & 0.3 & 0.4 & 0.5 & AVG \\
\hline
Random Baseline & 0.21 & 0.14 & 0.04 & 0.02 & 0.01 & 0.08 \\
Sultani et al.~\citeyearpar{sultani2018real} & 5.73 & 4.41 & 2.69 & 1.93 & 1.44 & 3.24 \\
AVVD~\citeyearpar{wu2022weakly} & 10.27 & 7.01 & 6.25 & 3.42 & 3.29 & 6.05 \\
VadCLIP~\citeyearpar{wu2024vadclip} & 11.72 & 7.83 & 6.40 & 4.53 & 2.93 & 6.68 \\
ITC~\citeyearpar{liu2024injecting} &13.54 &9.24 &7.45& 5.46&3.79 &7.90\\
ReFLIP~\citeyearpar{dev2024reflip} &14.23& 10.34& 9.32& 7.54&  6.81 & 9.62\\
Ex-VAD~\citeyearpar{huangex} & 16.51 & 12.35 & 9.41 & 7.82& 4.65 & \textbf{10.15}\\
\textbf{RefineVAD (Ours)} & 20.90 & 13.17 & 8.14 & 4.41 & 3.03 & \underline{9.93} \\
\bottomrule
\end{tabular}
\caption{\fontsize{10}{12}\selectfont Fine-grained comparisons on UCF-Crime.}
\label{tab:fine-ucf}
}
\end{table}

\begin{table}[t]
\centering
\resizebox{1.0\linewidth}{!}{
\small
\begin{tabular}{cccc}
\toprule
\multirow{2}{*}{MoTAR}   & \multicolumn{2}{c}{CORE} & \multirow{2}{*}{AUC(\%)} \\
\cline{2-3}
& Category-Injection & Soft-Classfication &  \\
\hline
&                         &                     & 84.60 \\
\checkmark &                         &                     & 85.43 \\
& \checkmark              &                     & 87.28 \\
\checkmark & \checkmark              &                     & 87.85 \\
\checkmark & \checkmark              & \checkmark          & \textbf{88.89} \\
\bottomrule
\end{tabular}
}
\caption{\fontsize{10}{12}\selectfont Ablation studies of our module on the UCF-Crime.}
\label{tab:ablation}

\end{table}

\subsection{Implementation Details}

We divide each video into 32 snippets and use pre-trained CLIP(VIT-L/14)~\cite{joo2023clip} as both the image and text encoder. For textual representation, InternVideo2.5~\cite{wang2025internvideo25empoweringvideomllms} is utilized. The encoders are kept frozen during training. For the XD-Violence dataset, we train the model for a maximum of 30 epochs using the AdamW optimizer (LR: 3e-3, Batch Size: 64), where each mini-batch consisted of 32 normal and 32 abnormal samples. The loss weights were set to $\lambda_1=0.1$ and $\lambda_2=0.2$. The implementation used PyTorch on a single NVIDIA A5000 GPU.

\subsection{Comparison on WVAD Benchmarks}

We compare our performance with current state-of-the-art (SOTA) methods on UCF-Crime and XD-Violence datasets. According to Table~\ref{tab:comparision_of_sota}, our model achieves 88.92\% AUC on UCF-Crime and surpasses all previous weakly supervised learning models on XD-Violence with 88.66\% AP. Regarding the segment-level evaluation in Table~\ref{tab:fine-ucf}, our method achieves competitive mAP across IoU thresholds on UCF-Crime, ranking among the top existing approaches. Unlike prior works that explicitly optimize snippet-level classification, our framework predicts only a single video-level category, which is then used to guide temporal localization. While this design improves category-level agreement, it places our model at a structural disadvantage under higher IoU thresholds, where the metric emphasizes frame-accurate boundary localization rather than alignment with video-level predictions. The slight performance gap at high IoU thus arises from this evaluation–objective mismatch rather than from the capability of our approach. On XD-Violence, our method remains competitive and surpasses previous weakly supervised methods by 0.19\%, validating the benefit of adaptive category-aware feature refinement via MoTAR and CORE.

\subsection{Ablation Studies}

To assess the contribution of each component, we conduct an ablation study on UCF-Crime (Table~\ref{tab:ablation}). Starting from the Base model (MLP with MIL), we sequentially add MoTAR, Category-Injection, and Soft Classification.

Introducing MoTAR provides a modest AUC gain (from 84.60\% to 85.43\%), showing the utility of motion-aware temporal adjustment. Notably, Category-Injection brings a much larger improvement, raising the AUC to 87.28\% and delivering the most significant single-step boost. Finally, adding Soft Classification—where embeddings are formed not from a single top category but from a weighted sum over category probabilities—further refines the representation. Combining all components yields the best performance at 88.89\%, confirming that each module is effective and that their integration provides complementary benefits for weakly supervised video anomaly detection.

\subsection{Cross-Dataset Semantic Transfer Evaluation}




To evaluate the transferability of the learned category semantics, we perform cross-dataset experiments on XD-Violence, as summarized in Table~\ref{tab:cross_dataset_transfer}. Full training on XD-Violence serves as the upper-bound reference, achieving 88.66\% AP. When freezing the CORE module (category classifier and semantic embeddings learned solely from UCF-Crime) and training the remaining modules, the model still attains 87.52\% AP, indicating the UCF-Crime semantic space effectively encodes transferable category-level structure despite dataset differences. Finally, a strict zero-shot test—evaluating a UCF-Crime trained model directly on XD-Violence—achieves a robust 77.56\% AP, demonstrating that the prototype-based semantics retain discriminative power across domains with mismatched category taxonomies and distinct visual distributions.

Collectively, these results show that our semantic representations extend beyond dataset-specific labels. Unlike prior category-driven approaches that required collecting new annotations for each target dataset and often degraded under mismatch, our compositional semantic space is label-efficient, scalable, and reusable without additional supervision, enabling practical cross-dataset deployment in weakly supervised video anomaly detection.

\newcommand{\cmark}{\ding{51}}
\newcommand{\xmark}{\ding{55}}
\begin{table}[t]
\centering
\resizebox{\linewidth}{!}{
\huge
\begin{tabular}{clcc}
\toprule
& Method & Category GT & AP (\%) \\
\midrule
& \makecell[l]{Full Training} 
    & \cmark & 88.66 \\
\addlinespace[4pt] 
& \makecell[l]{Freezing CORE Module \\\Large{(Category Classifier \& Embeddings: Trained only on UCF)}} 
    & \xmark & 87.52 \\
\addlinespace[4pt] 
& \makecell[l]{Direct Cross-Domain Transfer \\\Large{(Train: UCF-Crime, Test: XD-Violence, No Fine-Tuning)}} 
    & \xmark & 77.56 \\
\bottomrule
\end{tabular}
}
\caption{\fontsize{10}{12}\selectfont Semantic transfer of UCF-Crime category classifier and embedding on XD-Violence.}
\label{tab:cross_dataset_transfer}
\end{table}

\begin{figure}[t]
\centering
\includegraphics[width=1.0\columnwidth]{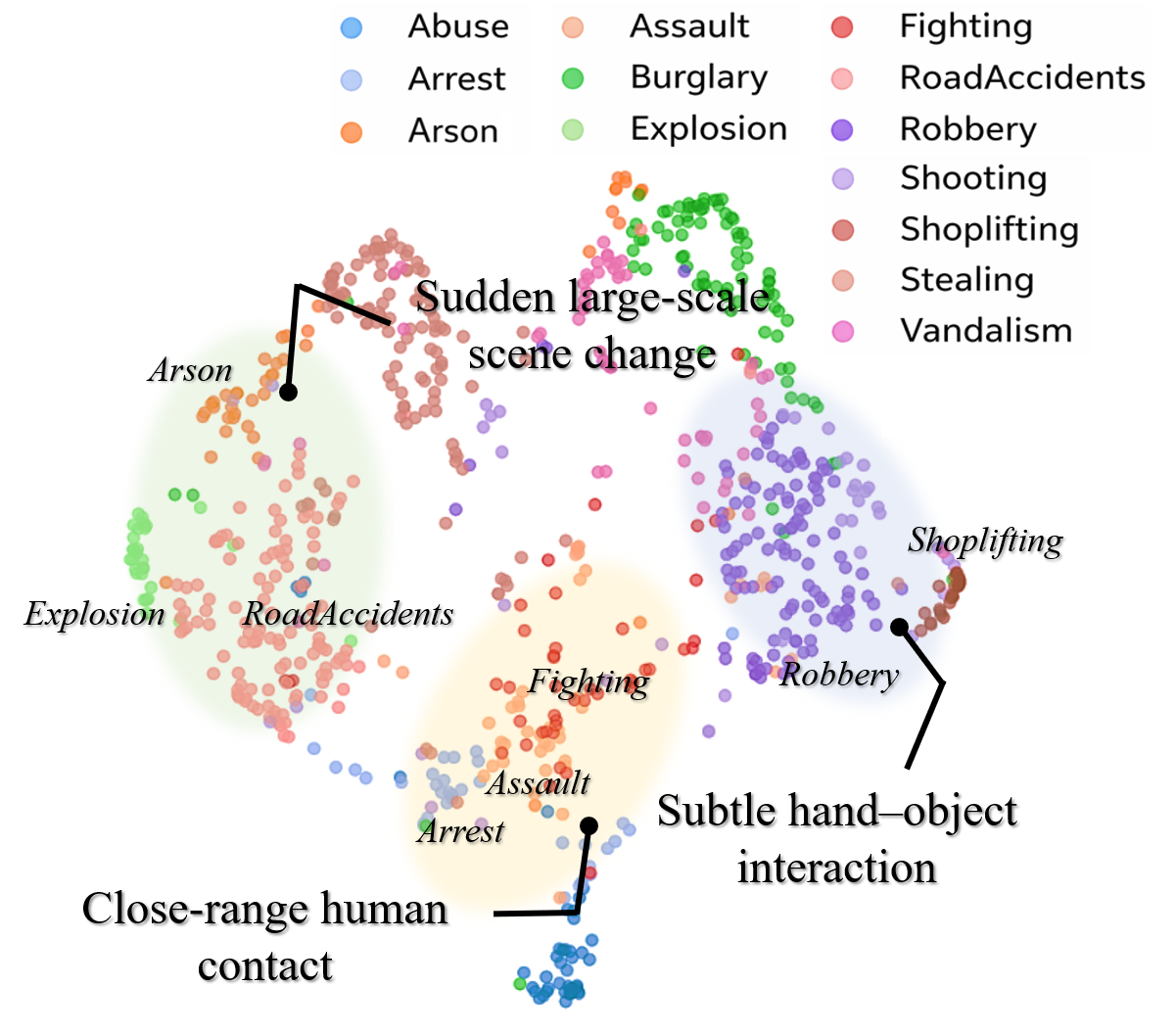} 
\caption{t-SNE visualization of logit features, where semantically similar categories form meaningful clusters.}
\label{fig4}
\end{figure}

\begin{figure}[t]
\centering
\includegraphics[width=1.0\columnwidth]{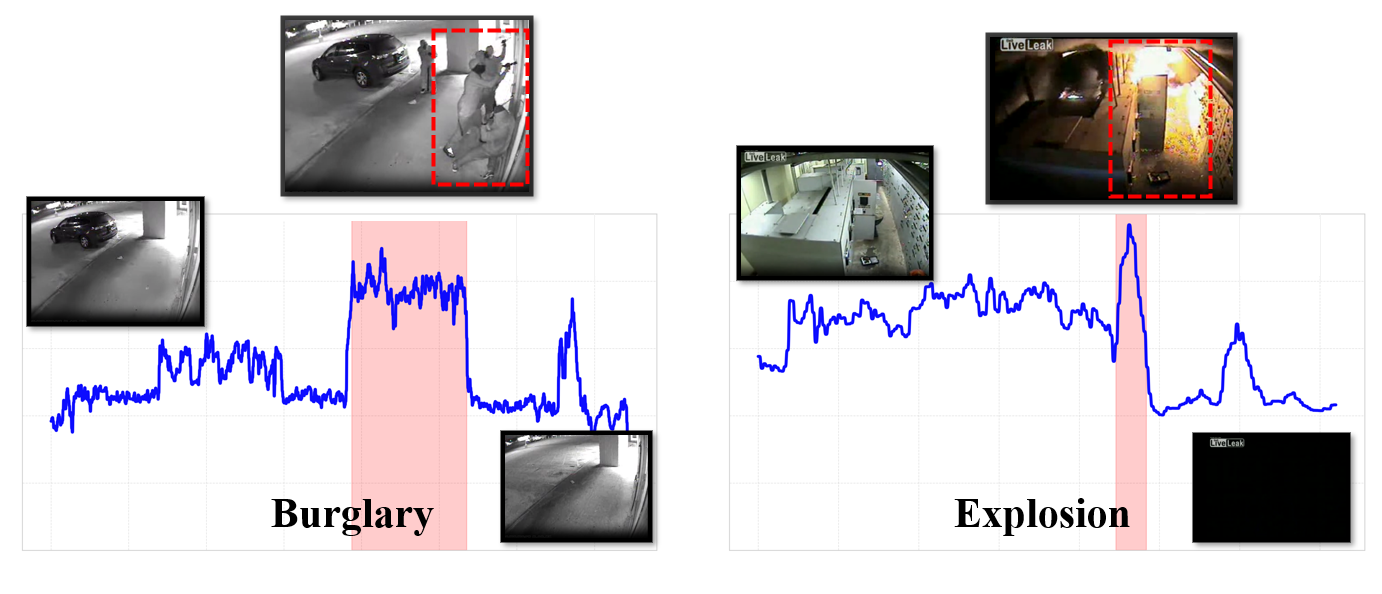} 
\caption{Example frame (top) and corresponding predicted scores with ground truth (bottom). The blue curve represents the predicted scores, and the red shaded regions indicate the ground-truth anomalous intervals.}
\label{fig5}
\end{figure}

\subsection{Category Logits Visualization}
To examine whether the predicted logit vectors are effectively separated according to semantic categories, we visualize the outputs of our soft category classifier using t-SNE. The logit vector corresponds to the category-wise activation output from the classifier, and similar vectors tend to be positioned close together in the embedding space.

As shown in the Figure~\ref{fig4}, samples with similar semantic characteristics form distinct clusters. For instance, Arson, Explosion, and RoadAccidents are grouped together, where the classifier tends to focus on sudden background changes or scene-level phenomena. In contrast, Arrest, Assault, and Fighting are clustered with a focus on the interactions among multiple people. On the other hand, Shoplifting and Robbery form a third group, where the model mainly attends to the actions of a single individual.

This observation suggests that different anomaly categories require attention to different aspects of the scene (e.g., background changes, group activity, or individual behavior), with similar categories exhibiting similar logit activation distributions. These results validate the effectiveness of category-aware anomaly detection, where adapting the decision-making process based on the anomaly type leads to more accurate detection.

\section{Conclusion}

In this work, we proposed RefineVAD, a weakly-supervised framework that integrates motion-aware temporal recalibration and category-conditioned feature refinement to enhance anomaly localization. By jointly modeling motion patterns and semantic categories, RefineVAD captures both generic and category-specific anomaly cues, enabling more precise detection. Our soft classification mechanism guides the model toward meaningful temporal and semantic evidence, leading to improved performance across benchmarks. In addition, the resulting category-aware feature space provides improved interpretability and demonstrates strong transferability across datasets. This approach also lays a foundation for future extensions toward fine-grained anomaly taxonomy, adaptive prototype refinement, and label-efficient cross-domain generalization.

\section{Acknowledgments}
This work was supported by the National Research Foundation of Korea (NRF) grant funded by the Korea government(MSIT)(RS-2024-00456589) and Korea Planning \& Evaluation Institute of Industrial Technology(KEIT) grant funded by the Korea government(MOTIE) (No. RS-2024-00442120, Development of AI technology capable of robustly recognizing abnormal and dangerous situations and behaviors during night and bad weather conditions)

\bibliography{aaai2026}

\end{document}